\documentclass[conference]{IEEEtran}
\IEEEoverridecommandlockouts
\usepackage{cite}
\usepackage{amsmath,amssymb,amsfonts}
\usepackage{algorithmic}
\usepackage{graphicx}
\usepackage{subfig}
\usepackage{textcomp}
\usepackage{xcolor}
\usepackage{colortbl}%
\usepackage{multirow}
\usepackage{colortbl}%
\usepackage{booktabs}
\usepackage{hyperref}
\usepackage{bbm}
  
\newcolumntype{C}[1]{>{\centering\arraybackslash}m{#1}}

\def\BibTeX{{\rm B\kern-.05em{\sc i\kern-.025em b}\kern-.08em
    T\kern-.1667em\lower.7ex\hbox{E}\kern-.125emX}}

\begin{document}

\title{SenseCF: LLM-Prompted Counterfactuals for Intervention and Sensor Data Augmentation}

\author{Shovito Barua Soumma$^{1,2}$, Asiful Arefeen$^{1,2}$, Stephanie M. Carpenter$^{1}$, Melanie Hingle$^{3}$ and Hassan Ghasemzadeh$^{1}$\vspace{-2mm}
\thanks{
$^{1}$College of Health Solutions, Arizona State University, Phoenix, AZ 85004, USA. Email: \{shovito, aarefeen, stephanie.m.carpenter, hghasemz\}@asu.edu.}
\thanks{
$^{2}$School of Computing and Augmented Intelligence, Arizona State University, Tempe, AZ 85281, USA.}
\thanks{$^{3}$School of Nutritional Sciences and Wellness, University of Arizona, Tucson, AZ, USA. Email: hinglem@arizona.edu}%
\thanks{This work was supported in part by the National Science Foundation (NSF) under grant IIS-2402650 and A. Arefeen is supported by the National Institute of Diabetes and Digestive and Kidney Diseases of the National Institutes of Health (NIH) under Award T32DK137525. The content is solely the responsibility of the authors and does not necessarily represent the official views of the NSF and NIH.}
\vspace{-2mm}
}

\maketitle

\begin{abstract}
Counterfactual explanations (CFs) offer human-centric insights into machine learning predictions by highlighting minimal changes required to alter an outcome. Therefore, CFs can be used as (i) interventions for abnormality prevention and (ii) augmented data for training robust models. In this work, we explore large language models (LLMs), specifically GPT-4o-mini, for generating CFs in a zero-shot and three-shot setting. We evaluate our approach on two datasets: the AI-Readi flagship dataset for stress prediction and a public dataset for heart disease detection. 
Compared to traditional methods such as DiCE, CFNOW, and NICE, our few-shot LLM-based approach achieves high plausibility (up to 99\%), strong validity (up to 0.99), and competitive sparsity.
% We compare LLM-generated CFs against state-of-the-art methods such as DICE, CFNOW, and NICE, assessing their plausibility, sparsity, validity, and impact on a downstream model's performance. 
% Our results show that LLM-based CFs, especially with few-shot prompting, generate highly plausible (up to 99\%) and sparse explanations that rival traditional optimization-based techniques.
Moreover, using LLM-generated CFs as augmented samples improves downstream classifier performance (an average accuracy gain of 5\%), especially in low-data regimes. This demonstrates the potential of prompt-based generative techniques to enhance explainability and robustness in clinical and physiological prediction tasks. Code base: 
\href{https://github.com/shovito66/SenseCF}{\textcolor{blue}{github.com/shovito66/SenseCF}}.
\end{abstract}


\vspace{-1.5mm}
\section{Introduction and Related Work}
\vspace{-1mm}

Accurate and interpretable predictions from machine learning (ML) models are increasingly vital in healthcare applications such as disease risk forecasting and sleep efficiency estimation using physiological and sensor data. While these models excel at outcome prediction, they often fall short in guiding actionable interventions to reverse adverse outcomes—especially in black-box settings.

Counterfactual explanations (CFs) offer a powerful solution by revealing the minimal changes needed to flip a model’s prediction. Traditional CFE methods like DiCE, CFNOW, and NICE rely on optimization procedures that often require model internals or gradient access, limiting their real-world applicability and struggling with categorical plausibility. In contrast, large language models (LLMs) provide a promising alternative: leveraging zero- and few-shot prompting, they can generate realistic, coherent counterfactuals using only input-output context~\cite{mann2020language}. This paradigm not only removes the dependence on gradients or model access but also opens the door for scalable, interpretable explanations across diverse datasets.

Recent work highlights LLMs’ innate counterfactual reasoning capabilities without fine-tuning~\cite{fizle_huan,li-etal-2024-prompting}, yet their use in structured, multimodal health data remains underexplored. Importantly, CFs not only enhance interpretability but can also serve as data augmenters, introducing label-flipping samples to strengthen models—particularly in imbalanced medical datasets. For data augmentation, CFs enhance robustness by introducing label-flipping variations while preserving data distributions. Optimization methods show promise in medical or low-data contexts but struggle with categorical coherence—a gap addressed by LLMs' semantic understanding~\cite{
% mazzine2023novel,
Brughmans2021NICEAA}.

However, several critical gaps persist in the current literature: first, the effectiveness of LLM-based CFs has not been comprehensively evaluated on multimodal clinical datasets; second, standardized evaluation metrics comparing optimization-based and generative approaches remain limited; third, CFs' potential as data augmenters in healthcare scenarios remains underexplored. 

% To address these gaps, our study systematically benchmarks LLM-generated counterfactuals against state-of-the-art methods (DICE, CFNOW, NICE) across heterogeneous healthcare datasets. Our work rigorously quantifies the diversity, plausibility, and augmentation efficacy of LLM-generated counterfactuals, highlighting their potential as both interpretability tools and effective data augmenters in multimodal clinical contexts. 

% In our study, we rigorously benchmark the LLM-generated counterfactuals against SOTA baselines on two clinically relevant multimodal datasets and show their potential to serve as both interpretative tools and effective data augmenters.

To address this gap, we introduce a systematic evaluation of zero- and few-shot LLM-generated counterfactuals across two real-world clinical datasets. Our contributions extend beyond existing LLM-focused studies that primarily evaluate natural language processing (NLP) tasks, providing a rigorous and quantitative comparison in multimodal clinical settings~\cite{fizle_huan,li-etal-2024-prompting}. We benchmark their plausibility, diversity, and impact on model performance against state-of-the-art baselines. To the best of our knowledge, this is the first study to explore LLMs as counterfactual generators for both explanation and augmentation in sensor-driven health contexts, moving toward AI systems that can inform not just prediction, but intervention.

\begin{figure}[!htb]
\vspace{-6mm}
\centering
\includegraphics[width=0.8\linewidth]{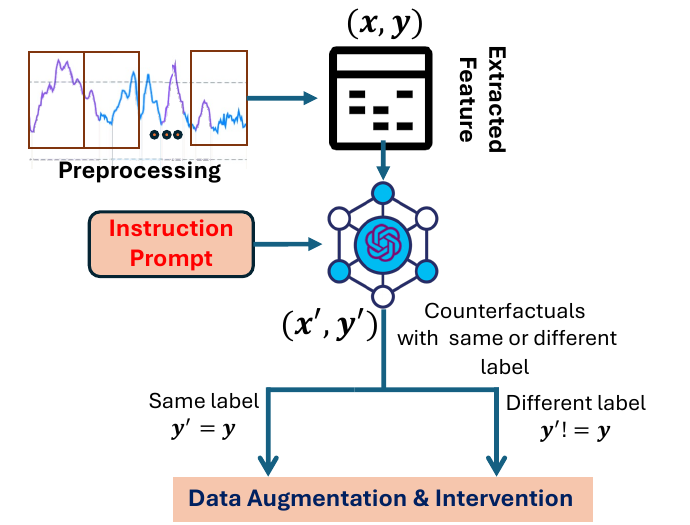}
\vspace{-2mm}
\caption{Counterfactual generation using LLMs from sensor-derived features.}
\label{fig:llm-cf-pipeline}
\vspace{-7mm}
\end{figure}
\vspace{-3mm}
\section{Methods}
\begin{figure*}[!htb]
\vspace{-3mm}
\centering
\includegraphics[width=0.8\linewidth]{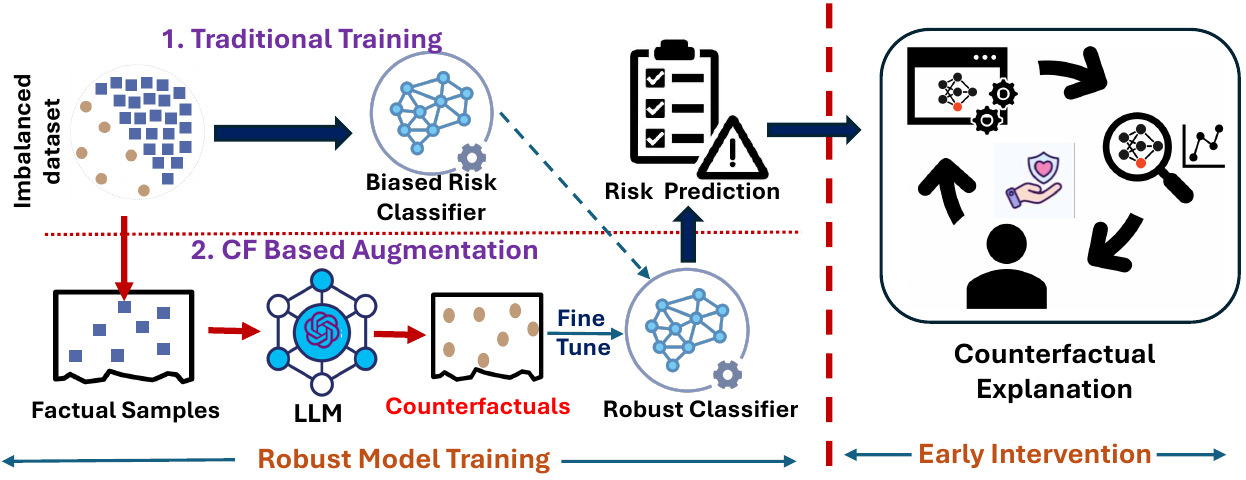} 
\vspace{-3mm}
\caption{SenseCF pipeline: LLM-generated counterfactuals are used both for augmenting imbalanced training data (left) and for model interpretability (right).}
\label{fig:method}
\vspace{-6mm}
\end{figure*}

In this section, we detail our approach for generating CFs using GPT-4o, structured as illustrated in Fig~\ref{fig:llm-cf-pipeline} and Fig~\ref{fig:method}. Our methodology aims at: (1) producing actionable counterfactual (CF) interventions by reversing the predictions of trained ML models, and (2) leveraging these CFs as augmented training data to enhance model performance, specifically addressing potential data imbalance.
We represent our input data as a set of tuples $(x_i,y_i)$ , where $x_i\in X$ is a feature vector representing either clinical or physiological data and $y_i\in \{0,1\}$ denotes the ground truth label. The trained predictive model $f(\cdot)$ outputs predictions $\hat{y_i}=f(x_i)$. Our preprocessing stage transforms raw data into structured, tabular feature vectors suitable for prompting the LLM. 

\subsection{Counterfactual Generation}
We used GPT-4o as an off-the-shelf counterfactual generator using specifically crafted instruction prompts in both zero-shot and few-shot settings. Formally, given a feature vector $x_i$ and the prediction $\hat{y_i}$, GPT-4o generates a modified vector $x^{'}_i$, where the model’s prediction changes from $\hat{y_i}$ to a desired opposite outcome $y_i\neq\hat{y_i}$. We also explicitly constrain the LLM from altering immutable or clinically fixed features (e.g., age, sex, or medication type), ensuring that generated counterfactuals remain actionable and plausible within domain constraints. The generation of CFs can be described as:
\vspace{-1mm}
\[
x'_i = \text{LLM}(x_i, \text{prompt}), \quad \text{subject to } f(x'_i) \neq f(x_i)
\vspace{-1.2mm}
\]
The instructional prompt explicitly constrains GPT-4o to minimally alter feature values to achieve a realistic and actionable counterfactual, ensuring the plausibility and feasibility of generated CFs.
\begin{figure}[!htb]
\vspace{-2mm}
\centering
\includegraphics[width=0.92\linewidth]{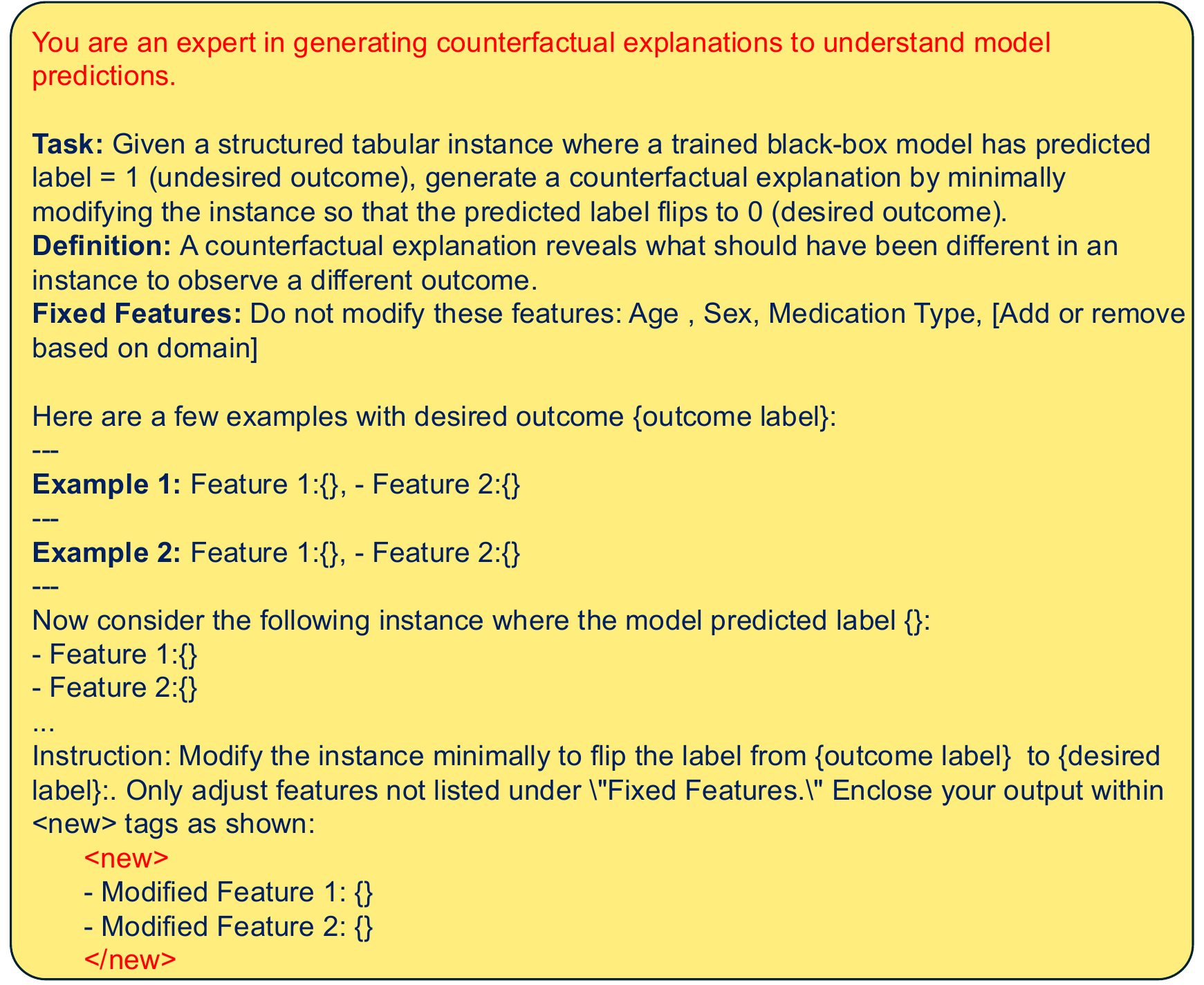}
\vspace{-1.6mm}
\caption{Prompt template for counterfactual generation.}
\label{fig:llm-prompt}
\vspace{-3mm}
\end{figure}
\subsection{Intervention and Data Augmentation}
Generated CFs serve dual purposes: (1) acting as plausible intervention points, and (2) augmenting training data. Specifically, CFs that successfully reverse predictions are included in the training dataset to enhance the robustness of the predictive model $f(\cdot)$. Formally, the augmented training set $X_{aug}$ is defined as:
\vspace{-2.5mm}
\[
X_{\text{aug}} = X \cup \{(x'_i, y'_i) \mid f(x'_i) \neq f(x_i)\}
\vspace{-1mm}
\]
Augmentation with CFs is particularly effective in handling imbalanced datasets, introducing valuable variability and balanced class representation, thus improving predictive model performance.
\vspace{-2mm}
\section{Experiment}
\vspace{-1.5mm}
\subsection{Data}
1. The \textbf{AI‑READI} data, part of an NIH-funded initiative, combines rich data sources—like vitals, ECG signals, continuous glucose monitoring (CGM), physical activity, and eye imaging—collected from 1,067 participants, including both diabetic and non-diabetic. In this analysis, we used twelve features to design counterfactual interventions for transitioning individuals from high stress ($y = 1$) to moderate stress levels ($y = 0$). Of the twelve features, we considered age, sex, and medication as immutable features; the rest are either raw or derived features from the sleep and daily glucose information collected using sensors (Garmin Vivosmart 5 and Dexcom G6).

2. Modified \textbf{Heart Disease} dataset from \cite{fedesoriano2021heart} has $918$ instances. We used four categorical and five continuous features to map whether an individual will have a heart disease $(y=1)$ or not $(y=0)$ and design interventions based on five features. A subset of the features, including fasting blood sugar level, resting blood pressure, resting ECG, and exercise-induced S-T depression, are derived using wearable sensors.
\vspace{-1mm}
\subsection{Baselines}
We have identified the following techniques to compare against SenseCF.

\textit{\textbf{DiCE}} \cite{mothilal2020dice} identifies a set of CFs by optimizing for proximity, diversity and sparsity. \textit{\textbf{CFNOW}} \cite{DEOLIVEIRA2023} searches an optimal point close to the factual point where the classification differs from the original. CFNOW performs greedy optimization for metrics like speed, coverage, distance, and sparsity. \textit{\textbf{NICE}}~\cite{Brughmans2021NICEAA} CFs are not necessarily adversarial data points but nearby instances in the data that reflect the desired outcome.

\subsection{Evaluation Metrics}
We assess the CFs using some standard metrics found in the literature:

\textbf{\textit{Validity}} assesses whether the produced CFs genuinely belong to the desired class. High validity indicates the technique’s effectiveness in generating valid CF examples.
\begin{equation*}
\vspace{-1.5mm}
    \textit{validity} = \frac{\#|f(X_T^*) \neq f(X_T)|}{\|CF\|}
\vspace{-1.5mm}
\end{equation*}

\textbf{\textit{Distance}} between the CF and the factual sample is calculated from the $L_2$ normalized distance of the continuous features and the hamming distance of the categorical features.

\textbf{\textit{Sparsity}} is the average number of feature changes per CF. A low sparsity ensures better user understanding of the CFs.
\begin{equation}
\vspace{-3mm}
\textit{sparsity} = \frac{\sum_{X_T^*\in CF}^{}\sum_{i=1}^{d} \mathbbm{1}(x_T^{*i} \neq x_T^i)}{\|CF\|} 
\vspace{-1mm}
\end{equation}

\textbf{\textit{Plausibility}} quantifies the fraction of explanations that fall within the feature ranges derived from the data-
\begin{equation*}
\vspace{-1mm}
    \textit{plausibility}=\frac{\sum_{X_T^*\in CF}^{}\mathbbm{1}(\text{dist}(X_T^*)\subseteq\text{dist}(X))}{\|CF\|}
\vspace{-1.5mm}
\end{equation*}
where, dist($X_T^*$) and dist($X$) represent the distribution of feature values in the CF instances $X_T^*$ and in the training data, respectively. $\|CF\|$ is the total number of CF instances.

\vspace{-2mm}
\section{Results}
Our evaluation highlights the dual role of LLM-generated CFs—as highly plausible interventions and as impactful data augmenters for robust model training in digital health contexts.
\vspace{-1.8mm}
\subsection{Intervention}
% \vspace{-2.5mm}
\begin{table}[h]
    \small    
    \centering
    \setlength{\tabcolsep}{0.6pt} % Adjust the horizontal padding between columns

    {\renewcommand{\arraystretch}{0.4}
    \begin{tabular}{p{1.4in}!{\vrule width 1pt}p{2in}}
         \toprule
         \cellcolor{red!25}\text{\textbf{Condition}} & \cellcolor{green!25}\text{\textbf{Intervention}}  \\
         \midrule
        A 81-year-old patient labeled as ``\textcolor{red}{stressed}'' showed low deep sleep (30.1\%), moderate REM (15.4\%), high blood glucose (210.8 mg/dL), and limited activity (5.95 steps). Stress level was high (85.25), with poor glucose control (TIR: 12.5\%, 1 hyper event). & 
        The LLM suggests \textit{\textbf{increasing deep sleep}} to \textcolor{blue}{$\uparrow$}35\% and \textit{\textbf{REM sleep}} to \textcolor{blue}{$\uparrow$}20\%, which could help reduce physiological and emotional stress. It also recommends \textit{\textbf{lowering blood glucose}} from 210.8 to \textcolor{red}{$\downarrow$}180 mg/dL, aligning with better metabolic control. These changes reflect clinically actionable strategies such as sleep hygiene improvement and tighter glucose management.\\
        \bottomrule
    \end{tabular}
    }
    \caption{Example of LLM-suggested counterfactual intervention for a high-stress patient}
    \label{tab:cf_intervention}
    \vspace{-5mm}
\end{table}

\begin{table}[h]
\vspace{-3mm}
\small
\caption{Evaluating the CFs on AI-READI Dataset.}
\vspace{-1.5mm}
\label{aireadi_result}
\centering
{\renewcommand{\arraystretch}{0.6}
% \scalebox{0.9}{
\begin{tabular}{p{0.5in}!{\vrule width 1pt}C{0.475in}C{0.48in}C{0.48in}C{0.65in}}
\toprule
 % \multirow{2}{*}{Method} &\multicolumn{4}{c}{\textbf{Validation metrics}} \\
Method & \cellcolor{blue!18}\text{validity $\uparrow$} &  \cellcolor{red!25}\text{distance $\downarrow$} & \cellcolor{green!25}\text{sparsity $\downarrow$} &  \cellcolor{gray!18}\text{plausibility $\uparrow$} \\
\midrule 
Zero-shot &\color{blue}0.91  &1.1  &3.6  &85      \\
3-shot  &\color{red}0.99  &1.2  &4.4  &\color{blue}99      \\
\midrule
DiCE     &0.67  &0.2  &2.27  &\color{red}100      \\
NICE     &0.85  &0.1  &2.9  &\color{red}100      \\
CFNOW    &0.44  &\color{red}0.02 &\color{red}1.12 &33      \\
\bottomrule
\end{tabular}
}
\vspace{-2mm}
% }
\end{table}
\begin{table}[h]
\small
\caption{Evaluating the CFs on Heart Disease Dataset.}
\vspace{-2.5mm}
\label{hr_result}
\centering
{\renewcommand{\arraystretch}{0.6}
% \scalebox{0.9}{
\begin{tabular}{p{0.5in}!{\vrule width 1pt}C{0.475in}C{0.48in}C{0.48in}C{0.65in}}
\toprule
 % \multirow{2}{*}{Method} &\multicolumn{4}{c}{\textbf{Validation metrics}} \\
Method & \cellcolor{blue!18}\text{validity $\uparrow$} &  \cellcolor{red!25}\text{distance $\downarrow$} & \cellcolor{green!25}\text{sparsity $\downarrow$} &  \cellcolor{gray!18}\text{plausibility $\uparrow$} \\
\midrule 
Zero-shot &0.88  &4.2  &7.9  &97.1      \\
3-shot &\color{red}0.97  &2.6  &5.2  &\color{blue}98.1      \\
\midrule
DiCE     &0.60  &0.2  &2.3  &\color{red}100      \\
NICE     &0.85  &0.07  &2.5  &\color{red}100      \\
CFNOW    &0.44  &\color{red}0.02  &\color{red}1.13  &33     \\
\bottomrule
\end{tabular}
% }
\vspace{-2mm}
}
\vspace{-4.5mm}
\end{table}
A representative counterfactual intervention generated by the SenseCF is illustrated in Table~\ref{tab:cf_intervention}. As shown in Tables~\ref{aireadi_result} and~\ref{hr_result}, CFs generated by GPT-4o using zero-shot and few-shot prompting achieve consistently high validity scores (up to 0.99), while maintaining competitive sparsity and distance. Notably, few-shot prompting improves realism and interpretability (e.g., 99\% plausibility on AI-READI, 98.1\% on HR), underscoring the LLM's semantic alignment with the target domain. Unlike optimization-based methods (e.g., DICE, CFNOW), which rely on access to model internals, our LLM-based approach generates CFs in a model-agnostic fashion while remaining interpretable and actionable. 

\begin{figure}[!h]
\vspace{-2.5mm}
\centering
\includegraphics[width=0.93\linewidth]{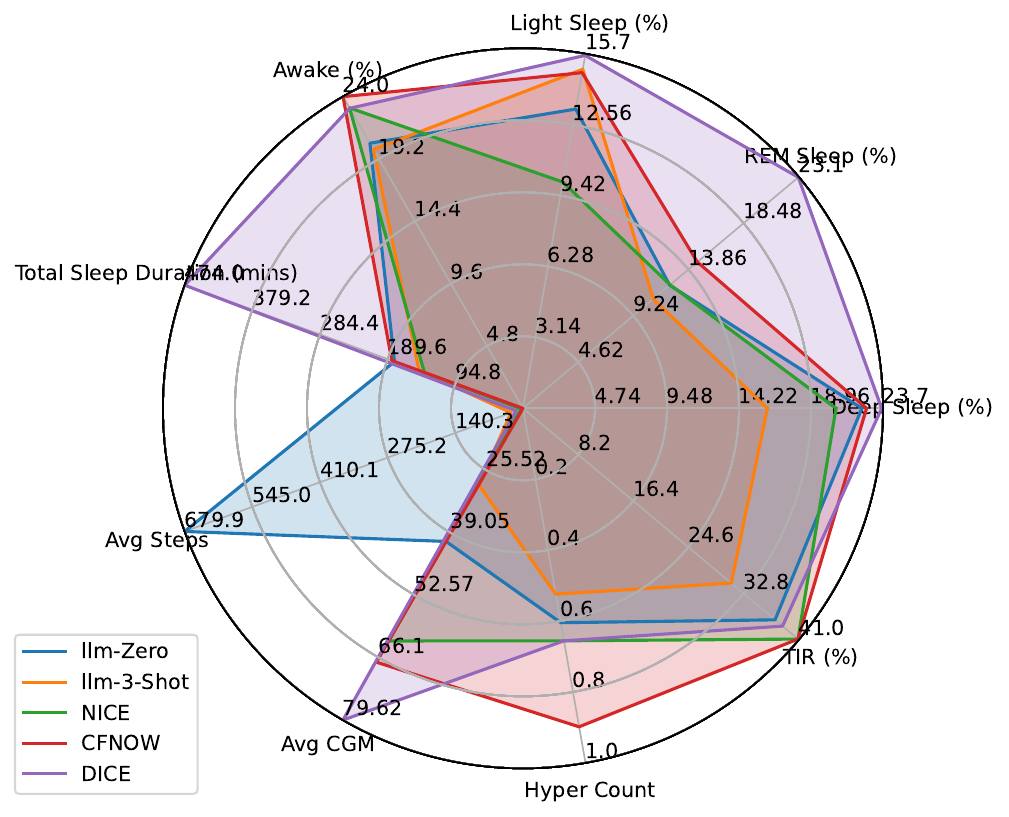}
\vspace{-1.6mm}
\caption{Feature diversity in the generated CFs for AI-Readi data. Avg: Average, Hyper: No of hyperglycemia}
\label{fig:diversity}
\vspace{-4mm}
\end{figure}

Figure~\ref{fig:diversity} shows the diversity of the CFs from different methods across all the mutable features. SenseCF-generated CFs exhibit the least diversity for all features, except that the 3-shot variant shows higher diversity in Average Step counts.

\subsection{Impact as Data Augmenters}
Beyond explanation, Tables~\ref{aireadi_augmentation} and~\ref{hr_augmentation} demonstrate that LLM-generated CFs significantly enhance model performance when used for data augmentation. Across classifiers—including Random Forest (RF), SVC, XGB, and NN—we observe performance improvements in terms of accuracy, precision, recall, and AUC. While the original models already perform well on balanced datasets (e.g., HR), the gains are especially relevant in realistic healthcare scenarios with data imbalance, as reflected in AI-READI outcomes. For instance, RF performance on AI-READI improves from 0.74 to 0.80 in accuracy with counterfactual augmentation.

The few-shot LLM prompts, in particular, lead to consistently high classification scores that match or exceed those obtained from traditional SOTA augmentation techniques like DICE, NICE, and CFNOW. This means that LLMs capture not only the structure of the data but also the semantic boundaries between classes—resulting in augmentations that are both label-flipping and label-preserving depending on the use case.

\begin{table}[!h]
\vspace{-2mm}
\footnotesize
% \caption{Impact of LLM and SOTA Counterfactual Augmentation on AI-Readi Model Performance}
\caption{Performance Impact of LLM-Generated and SOTA Counterfactuals on AI-Readi Classifiers}
\label{aireadi_augmentation}
\centering
{\renewcommand{\arraystretch}{0.5}
\begin{tabular}{c!{\vrule width 1pt}c|ccccc}
\toprule
\textbf{Model} &\textbf{Method} &\textbf{ACC} &\textbf{PRE} &\textbf{REC} &\textbf{F1} &\textbf{AUC}\\
\midrule
\multirow{7}{*}{RF} &Zero	&0.79	&0.79	&0.77	&0.78	&0.86\\
                    &Few	&0.793	&0.79	&0.78	&0.79	&0.87\\
                    &Both	&0.8	&0.8	&0.79	&0.79	&0.86\\
                    \cline{2-7}\rule{0pt}{2.2ex}
                    &DICE	&0.788	&0.79	&0.76	&0.78	&0.87\\
                    &NICE	&0.787	&0.78	&0.77	&0.78	&0.86\\
                    &CFNOW	&0.79	&0.78	&0.78	&0.78	&0.86\\
                    \cline{2-7}\rule{0pt}{2.2ex}
                &$\times$   &0.74	&0.72	&0.77	&0.74	&0.82\\
\midrule
\multirow{7}{*}{SVC}    &Zero	&0.63	&0.63	&0.57	&0.6	&0.67\\
                        &Few	&0.65	&0.64	&0.61	&0.62	&0.7\\
                        &Both	&0.65	&0.64	&0.6	&0.62	&0.7\\
                        \cline{2-7}\rule{0pt}{2.2ex}
                        &DICE	&0.65	&0.64	&0.6	&0.62	&0.7\\
                        &NICE	&0.65	&0.64	&0.6	&0.62	&0.7\\
                        &CFNOW	&0.65	&0.64	&0.61	&0.62	&0.7\\
                        \cline{2-7}\rule{0pt}{2.2ex} 
                     &$\times$  &0.63	&0.63	&0.58	&0.6   &0.67\\
\midrule
\multirow{7}{*}{XGB}    &Zero	&0.77	&0.77	&0.75	&0.76	&0.85\\
                        &Few	&0.78	&0.78	&0.75	&0.76	&0.86\\
                        &Both	&0.78	&0.78	&0.77	&0.77	&0.86\\
                        \cline{2-7}\rule{0pt}{2.2ex}
                        &DICE	&0.78	&0.79	&0.75	&0.77	&0.86\\
                        &NICE	&0.78	&0.79	&0.75	&0.77	&0.85\\
                        &CFNOW	&0.76	&0.77	&0.74	&0.76	&0.85\\
                        \cline{2-7}\rule{0pt}{2.2ex}
                        &$\times$  &0.74	&0.72	&0.77	&0.74	&0.81\\
\midrule
\multirow{7}{*}{NN}    &Zero   &0.64    &0.63	&0.6	&0.62   &0.67\\
                       &Few	&0.642	&0.64	&0.64	&0.63	&0.69\\
                       &Both	&0.633	&0.62	&0.62	&0.62	&0.67\\
                       \cmidrule(l){2-7}
                        &DICE	&0.635	&0.64	&0.56	&0.6	&0.68\\
                        &CFNOW	&0.64	&0.64	&0.57	&0.6	&0.69\\
                        &NICE	&0.638	&0.63	&0.63	&0.63	&0.69\\
                        \cline{2-7}\rule{0pt}{2.2ex}
                        &$\times$  &0.63	  &0.6	   &0.57   &0.58   &0.67\\
\bottomrule
\end{tabular}}
\vspace{-3mm}
\end{table}

\begin{table}[!h]
\footnotesize
% \caption{Impact of LLM and SOTA Counterfactual Augmentation on Heart Disease Model Performance}
\caption{Performance Impact of LLM and SOTA Counterfactual Augmentation on Heart Disease Classification Models}
\label{hr_augmentation}
\centering
{\renewcommand{\arraystretch}{0.5}
\begin{tabular}{c!{\vrule width 1pt}c|ccccc}
\toprule
\textbf{Model} &\textbf{Method} &\textbf{ACC} &\textbf{PRE} &\textbf{REC} &\textbf{F1} &\textbf{AUC}\\
\midrule
\multirow{7}{*}{RF} &Zero	&0.985	&1	&0.97	&0.99	&1\\
                    &Few	&0.985	&1	&0.97	&0.99	&1\\
                    &Both	&0.985	&1	&0.97	&0.99	&1\\
                    \cline{2-7}\rule{0pt}{2.2ex}
                    &DICE	&0.985	&1	&0.97	&0.99	&1\\
                    &NICE	&0.985	&1	&0.97	&0.99	&1\\
                    &CFNOW	&0.985	&1	&0.97	&0.99	&1\\
                    \cline{2-7}\rule{0pt}{2.2ex}
                    &$\times$  &0.9854	&1	&0.97	&0.99	&1\\
\midrule
\multirow{7}{*}{SVC}    &Zero	&0.79	&0.77	&0.83	&0.8	&0.88\\
                        &Few	&0.82	&0.78	&0.89	&0.83	&0.87\\
                        &Both	&0.8	&0.78	&0.83	&0.8	&0.88\\ 
                        \cline{2-7}\rule{0pt}{2.2ex}
                        &DICE	&0.81	&0.76	&0.9	&0.83	&0.87\\
                        &NICE	&0.82	&0.76	&0.93	&0.84	&0.87\\
                        &CFNOW	&0.80	&0.76	&0.87	&0.81	&0.87\\
                        \cline{2-7}\rule{0pt}{2.2ex}
                        &$\times$  &0.82	&0.76	&0.92	&0.83	&0.87\\
\midrule
\multirow{7}{*}{XGB}    &Zero   &0.985	&1	&0.97	&0.99	&0.99\\
                        &Few   &0.985	&1	&0.97	&0.99	&0.99\\
                        &Both   &0.985	&1	&0.97	&0.99	&0.99\\
                        \cline{2-7}\rule{0pt}{2.2ex}
                        &DICE	&0.91	&0.98	&0.83	&0.9	&0.96\\
                        &NICE	&0.91	&0.92	&0.87	&0.9	&0.95\\
                        &CFNOW	&0.91	&0.99	&0.83	&0.9	&0.97\\
                        \cline{2-7}\rule{0pt}{2.2ex}
                    &$\times$  &0.98	&0.98	&0.97	&0.98	&0.98\\
\midrule
\multirow{7}{*}{NN}     &Zero	&0.81	&0.78	&0.84	&0.81	&0.88\\
                        &Few	&0.78	&0.75	&0.84	&0.79	&0.88\\
                        &Both	&0.8	&0.79	&0.82	&0.81	&0.88\\
                        \cline{2-7}\rule{0pt}{2.2ex}
                        &DICE	&0.77	&0.75	&0.83	&0.78	&0.87\\
                        &CFNOW	&0.76	&0.74	&0.82	&0.77	&0.87\\
                        &NICE	&0.77	&0.74	&0.84	&0.79	&0.87\\
                        \cline{2-7}\rule{0pt}{2.2ex}
                        &$\times$  &0.78	&0.74	&0.85	&0.79	&0.87\\
\bottomrule

\end{tabular}}
\vspace{-5mm}
\end{table}

\section{Conclusion \& Future Work}
In this work, we introduce a novel framework for generating CFs using large language models (LLMs), with a focus on structured sensor-derived datasets in health and physiological monitoring. Our method leverages zero-shot and few-shot prompting with GPT-4o to generate semantically valid, plausible CFs that flip model predictions while respecting domain-specific constraints such as immutable features.

Through extensive evaluation on two real-world datasets, we show that LLM-generated CFs are not only effective for model interpretability and intervention design but also significantly enhance model performance when used for data augmentation, particularly in contexts affected by data imbalance. 
% Our approach is model-agnostic, requires no fine-tuning or gradient access, and is broadly applicable across clinical and wearable sensing domains. 
The method is model-agnostic, requires no gradient access, and generalizes across clinical and wearable domains. One limitation is the higher feature distance compared to baselines; future work will focus on refining prompts to improve counterfactual compactness and quality.
% \textcolor{red}{While our approach achieves strong plausibility and validity, one limitation is the relatively higher feature distance compared to baselines. Future work will explore prompt refinement and LLM control strategies to reduce distance and improve overall counterfactual quality.}

% To the best of our knowledge, this is the first work to systematically explore the use of LLMs for counterfactual generation in sensor-based data under both zero- and few-shot settings. We believe this opens a promising direction for integrating generative AI into interpretable and trustworthy healthcare machine learning pipelines.
To the best of our knowledge, this is the first systematic exploration of LLM-based CFs in sensor-driven data under both zero- and few-shot settings. We believe this opens a promising direction for integrating generative AI into trustworthy, intervention-oriented healthcare ML pipelines.

\vspace{-2mm}
\bibliographystyle{IEEEtran}
\bibliography{ref.bib}

\end{document}